\pdfoutput=1

\documentclass[11pt]{article}

\usepackage[final]{acl}

\usepackage{amsmath}
\usepackage{amssymb}
\usepackage{times}
\usepackage{latexsym}

\usepackage{enumitem} %
\usepackage{booktabs} %

\usepackage[T1]{fontenc}

\usepackage[utf8]{inputenc}

\usepackage{microtype}
\usepackage{textgreek}
\usepackage{inconsolata}

\usepackage{graphicx}
\usepackage{natbib}
\usepackage{subcaption}
\usepackage{authblk}

\newcommand{\ie}{\textit{i.e.}}
\newcommand{\eg}{\textit{e.g.}}

\newcommand{\update}[1]{\textcolor{black}{#1}}

\title{\textit{I see what you mean} \\
Co-Speech Gestures for Reference Resolution in Multimodal Dialogue}

\author[1,2]{\textbf{Esam Ghaleb}}
\author[3]{\textbf{Bulat Khaertdinov}}
\author[1,2]{\textbf{Asl\i~Özy\"{u}rek}}
\author[4]{\textbf{Raquel Fern\'andez}}

\affil[1]{Multimodal Language Department, Max Planck Institute for Psycholinguistics}
\affil[2]{Donders Institute for Brain, Cognition and Behaviour, Radboud University}
\affil[3]{Department of Advanced Computing Sciences, Maastricht University}
\affil[4]{Institute for Logic, Language and Computation, University of Amsterdam}
\affil{\small \textbf{Correspondence:} \href{mailto:esam.ghaleb@mpi.nl}{esam.ghaleb@mpi.nl}}

\begin{document}
\maketitle
\begin{abstract}
In face-to-face interaction, we use multiple modalities, including speech and gestures, to communicate information and resolve references to objects. However, how representational co-speech gestures refer to objects remains understudied from a computational perspective. In this work, we address this gap by introducing a multimodal reference resolution task centred on representational gestures, while simultaneously tackling the challenge of learning robust gesture embeddings. We propose a self-supervised pre-training approach to gesture representation learning that grounds body movements in spoken language. Our experiments show that the learned embeddings align with expert annotations and have significant predictive power. Moreover, reference resolution accuracy further improves when (1) using multimodal gesture representations, even when speech is unavailable at inference time, and (2) leveraging dialogue history. Overall, our findings highlight the complementary roles of gesture and speech in reference resolution, offering a step towards more naturalistic models of human-machine interaction.
\end{abstract}

\begin{figure}
    \centering
    \includegraphics[width=1\linewidth]{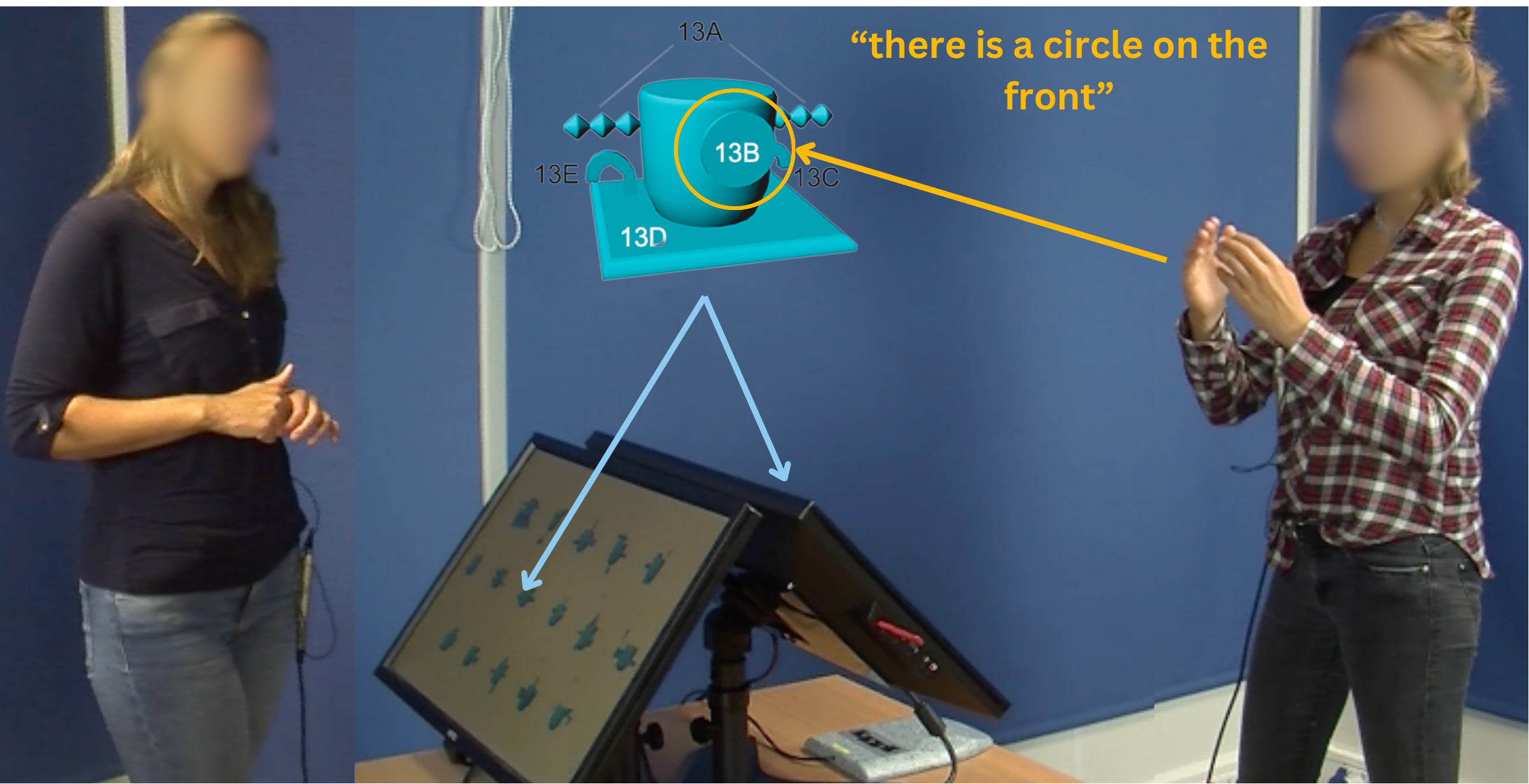}
    \caption{Example from the CABB dataset \cite{rasenberg2022primacy}, illustrating how participants resolve references through speech and gestures in face-to-face dialogue. The speaker on the right says ``there is a circle on the front'' while performing a representational 
    gesture that resembles a circle. 
    The object discussed is shown for illustration but not visible to the listener; the orange arrow points to the referent as annotated by experts. 
    Our work draws on these interactions to model \emph{multimodal reference resolution}.}
    \label{fig:example_interaction}
\end{figure}

\section{Introduction}

Referring to objects is common in everyday communication. 
In face-to-face interaction, when we need to collaborate on new tasks or refer to new objects, we rely on verbal (\ie, speech) and non-verbal (\eg, gestures and gaze) cues to describe salient object features and direct the listener's attention.
Among the non-verbal cues are \textit{representational co-speech gestures}, \ie, iconic hand movements semantically and pragmatically related to co-occurring speech \cite{kendon2004gesture}. 
Studies have shown that representational gestures facilitate language comprehension \cite{drijvers2017visual, arbona2023semantically} and help listeners identify referents more quickly than speech alone \cite{campana2005real}.
Along with speech, gestures are used to refer to novel objects and build shared understanding \cite{rasenberg2022primacy, Akamine2024sp}, as shown in Figure~\ref{fig:example_interaction}. 
These multimodal abilities are inherent ingredients of our communicative interactions \cite{ozyurek2014hearing}. Hence, developing computational approaches that can interpret such cues is important for naturalistic human-machine collaboration: in situated dialogue, an artificial agent must recognize multimodal inputs and speaker references to meet human needs \cite{moon2020situated, kontogiorgos2018multimodal}. 

However, how to computationally represent and interpret co-speech gestures remains an understudied problem, particularly within Natural Language Processing. Other gestural forms, such as deictic gestures, \ie, pointing \cite{gatt-paggio-2013-empirical, KenningtonSchlangen2017, chen2021yourefit, kontogiorgos2018multimodal} or beat gestures, \ie, rhythmic movements without semantic content \cite[e.g.,][]{sinclair2021linguistic, abzaliev2022towards}, have received some attention. 
In contrast, the challenges posed by iconic, representational gestures and their contribution to reference resolution have hardly been tackled from a data-driven perspective. 

In this paper, we propose an approach to learning embeddings for representational gestures that exploits not only the body movements that make up a gesture but also the semantically related speech that is typically produced simultaneously with it. 
We combine contemporary self-supervised learning techniques to train a Transformer-based gesture encoder and ground it in information from features extracted by text or speech large language models. Then, we test the effectiveness of the pre-trained gesture embeddings in the downstream task of reference resolution in face-to-face dialogue, showing that gestures---as learned by our proposed multimodal approach---have significant predictive power that complements the verbal modality. More concretely, we make the following contributions: 

\begin{itemize}[leftmargin=10pt, itemsep=-1pt, topsep=0pt] %

\item We propose three model architectures for gesture representation learning that exploit a version of the motion encoder DSTFormer \cite{zhu2023motionbert}, which we adapt to allow for the integration of speech through cross-modal attention.
\item We show that the resulting pre-trained gesture embeddings are aligned with expert knowledge present in manual annotations, clearly surpassing earlier approaches to gesture representation learning \cite{ghaleb2024learning}.
\item We introduce a novel multimodal reference resolution task and demonstrate that learning gesture representations by jointly exploiting body movements and the semantics of the concurrent speech results in more accurate models, even when speech is not available at prediction time. 
\item Our reference resolution experiments also show that leveraging dialogue history improves model prediction and that, when speech is present at test time, gestures provide complementary information that enhances reference resolution accuracy. 
\item Our experiments make use of the CABB dataset \cite{rasenberg2022primacy,eijk2022cabb}, collected by cognitive scientists. We make available the pre-processed data and the code to reproduce all our results via a public GitHub repository, providing valuable resources to the community.\footnote{\href{https://github.com/EsamGhaleb/MultimodalReferenceResolution}{https://github.com/EsamGhaleb/ReferenceResolution}}
\end{itemize}

\section{Related Work}
\label{sec:related}

\subsection{Learning Multimodal Representations}
Despite the importance of gestures in multimodal communication, learning gesture representations remains challenging and understudied in both computer vision and NLP.  Some existing work has used formal approaches to integrate gestures into discourse semantics \cite{lascarides2009formal,lai2024encoding}, while a few other works have employed data-driven methods. For example, \citet{abzaliev2022towards} jointly learned gesture and word embeddings from TED talks using contrastive learning, and showed that function words, discourse markers, and the language of the speaker can be predicted from non-representational gestures. 
Self-supervised contrastive learning techniques \cite{chen2020simple,radford2021learning} have been widely adopted in the field of multimedia to learn representations of human movements from skeletal joint coordinates unimodally \cite{thoker2021skeleton,zhu2023motionbert} and in combination with other data modalities \cite{brinzea2022contrastive, liu2024multi}, while \cite{lee2021crossmodal} used self-supervised learning to learn gesture embeddings as a pre-training stage for gesture generation.

Our approach to learning gesture representations is most closely related to the preliminary work of \citet{ghaleb2024learning}, who proposed to learn embeddings for representational gestures by grounding them in co-occurring speech. We substantially extend this work by replacing their skeleton encoder with a Transformer-based encoder, allowing us to integrate not only speech but also text-based semantic embeddings with higher temporal granularity and using a much larger amount of data samples. Furthermore, unlike this work, we exploit the learned gesture embeddings for the downstream task of reference resolution, here formulated as the problem of identifying the object referred to by a gesture in face-to-face dialogue.

\subsection{Reference Resolution in Dialogue}

Reference resolution in dialogue has mostly been modelled as the task of identifying the referent of text-based linguistic expressions, ignoring non-verbal cues. %
For example, \citet{skantze2022collie} proposed COLLIE, a continual learning method that adjusts language embeddings to accommodate new language use for new referents; in an earlier study,  \citet{shore2018using} found that leveraging dialogue history in the form of previous referring expressions improves model prediction, similarly to \citet{takmaz2020refer}. 
Resolving linguistic referring expressions in the visual
modality has also been studied in the field of computer vision thanks to datasets such as ReferIt~\cite{kazemzadeh2014referitgame}, Flicker30k Entities~\cite{plummer2015flickr30k},
and Visual Genome~\cite{krishna2017visual}, which map referring expressions to regions in an image.

In this work, we focus on reference resolution in face-to-face communication, where linguistic expressions interact with non-verbal signals like gestures. The large majority of work in this domain has been concerned with deictic pointing gestures. For instance, \citet{kennington2017simple} combined linguistic information with gaze and deictic gestures by treating them as separate resolution models and then fusing their predictions via interpolation. Similarly, \citet{kontogiorgos2018multimodal} used multisensory input in a collaborative assembly task to assess the contribution of various cues--such as eye gaze, head direction, and pointing gestures--to reference resolution. They found that deictic gestures, when combined with speech, reliably located objects, while gaze and head direction were only useful for approximating the general location of the intended object when paired with speech. More recently, within the computer vision community, \citet{chen2021yourefit} found that referential expressions were more discriminative when both visual context and pointing gestures were considered, compared to using visual context alone.

In this paper, we tackle reference resolution by means of iconic representational gestures rather than pointing, calling attention to the importance of modelling such gestures to identify objects in multimodal interaction.

\section{Data}
\label{sec:data}
For our study, we use the CABB dataset \cite{eijk2022cabb,rasenberg2022primacy}, which consists of face-to-face conversations in Dutch between two dialogue participants who play a reference game. The setup is shown in Figure \ref{fig:example_interaction}. The participants' task is to identify 16 objects without conventional names that are made up of different geometrical parts (see Appendix~\ref{app:objects}). Each dyad plays the game for six rounds, exchanging the roles of `director' (who describes one of the target objects) and `matcher' (who attempts to identify the director's intended referent among the 16 candidate objects displayed on a screen). 
The participants are free to communicate as they like, which elicits spontaneous speech and gestures. Speakers were video recorded from different angles and we make use of the semi-frontal views shown in Figure \ref{fig:example_interaction}, as well as the audio recordings.

We use two different subsets of this data, which we refer to as CABB-S and CABB-L, plus an extension of the latter which we call CABB-XL: 

\paragraph{CABB-S} \cite{rasenberg2022primacy}
consists of 19 dialogues by 38 individuals, corresponding to over 8 hours of recordings. The dataset includes manual speech transcriptions and manual segmentation of gesture strokes, 
with 4949 gesture segments in total. 
Approximately 97\% of these segments are accompanied by concurrent speech.
CABB-S also includes manual annotations of gesture strokes with 
two types of information:\footnote{For further details, see \citet{rasenberg2022primacy}.} 

\begin{itemize}[itemsep=1pt,topsep=1pt,left=0pt] %
    \item \textbf{Referent:} The object subpart referred to by the gesture. The candidate objects and their sub-parts are shown in  Appendix~\ref{app:objects}, Figure \ref{fig:all_object_and_subparts}.
\item \textbf{Form similarity:} 419 pairs of gestures with the same referent are annotated with five low-level binary features indicating whether two semantically related gestures are similar regarding shape, movement, rotation, position, and use of hands.
\end{itemize}

\noindent
We use CABB-S for evaluating our pre-training approach to gesture representation learning (Section~\ref{sec:eval_pretraining}) and for the experiments on reference resolution (Section~\ref{sec:ref_resolution}). 

\paragraph{CABB-L} \cite{eijk2022cabb}
contains an additional 49 dialogues by 98 different subjects, with about 42.5 hours of recordings. It is therefore much larger than CABB-S. Only 42 dialogues are manually transcribed and no manual annotations regarding gestures are present. To identify gestures, we use the segmentation model by \citet{ghaleb2024le}, 
\update{which has been shown to achieve a mean Average Precision (mAP) of 76\% on the CABB-S dataset. Applying this model to CABB-L results in 30k automatically segmented gestures.}\footnote{\update{Some qualitative results on segmentation performance can be found in Appendix~\ref{app:segmentation_qualitative_results}.}}

To increase the amount of data available for pre-training, we oversample 
by selecting 1-second time windows overlapping more than 50\% with the automatically segmented gestures. This results in approximately 400k data samples, which we refer to as \textbf{CABB-XL}. 
We use Whisper-X \cite{bain2022whisperx} to automatically generate speech transcriptions when manual transcriptions are unavailable. 83\% of the gestures are accompanied by speech. We use CABB-L and CABB-XL for pre-training the models introduced in Section~\ref{sec:models}.

\paragraph{Pre-processing}
To process body movements, we apply the procedure used by \citet{ghaleb2024learning} to CABB-S, CABB-L, and CABB-XL. Concretely, we sample 1-second time windows centered around each segmented gesture and use ViTPose \cite{xu2022vitpose} to extract skeletal information, \ie, 2D keypoint coordinates 
for 27 upper body and hand joints. 
\update{\citet{holler2019multimodal} showed that related speech is often produced before or after 
gesture strokes (the most meaning-bearing segment of a gesture) by a few hundred milliseconds. To account for this temporal asynchrony between speech and gestures, when processing }
the verbal modality we extract 2-second windows centered around the sampled gestures, 
and use both the raw speech and the transcriptions as described in the next section.

\section{Gesture Representation Learning}
\label{sec:models}

\begin{figure*}
    \centering
    \includegraphics[width=0.99\linewidth]{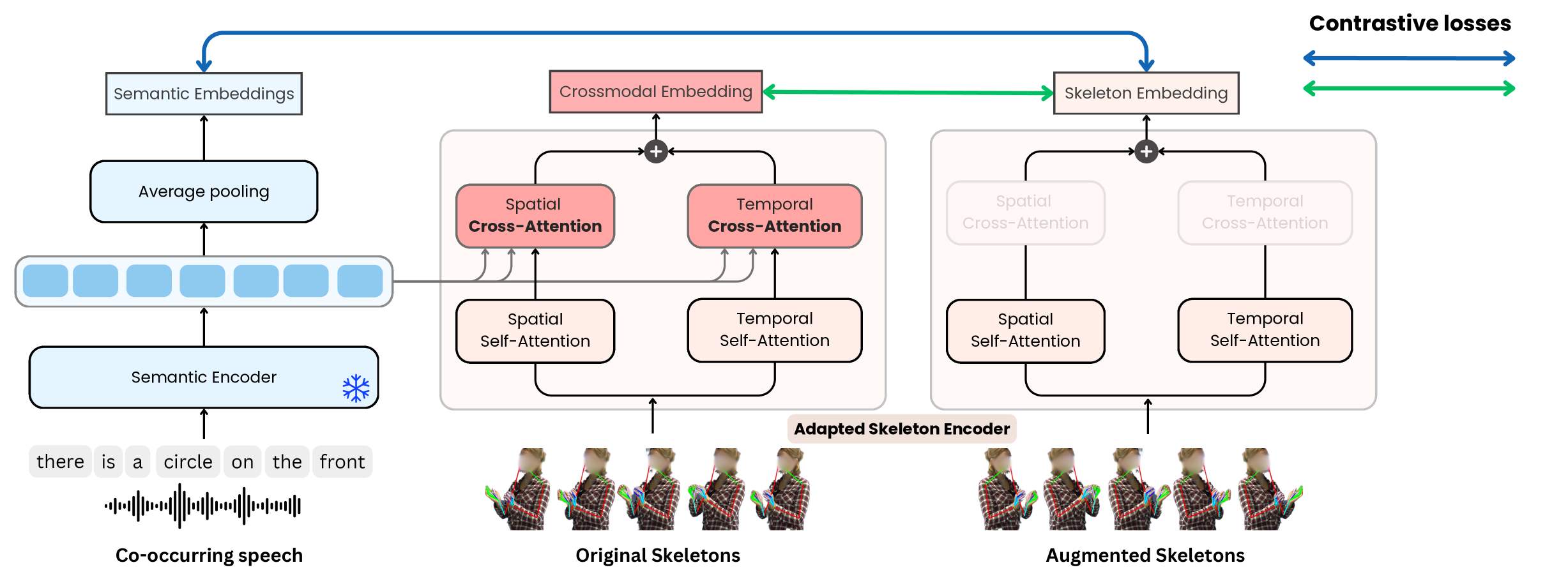}
    \caption{Our multimodal-X architecture. The left branch encodes semantic information (text or speech) and fuses it with skeleton embeddings via the proposed cross-attention blocks in our adapted skeleton encoder. The architecture is trained by minimizing contrastive losses.}
    \label{fig:multimodal-x_approach}
\end{figure*}

In this section, we present our approach to learning robust gesture representations in a self-supervised fashion. To do so in the context of \textit{multimodal} communication, we experiment with three types of input: gestures themselves (\ie, skeletal information corresponding to body movement), raw speech, and text-based semantics (Section~\ref{subsec:encoders}). We propose three pre-training model architectures that use these input types to different degrees and with different multimodal integration strategies (Section~\ref{sec:architectures}). We train these models on CABB-L/XL and evaluate them against expert annotations using the unseen gestures in CABB-S (Section~\ref{sec:eval_pretraining}). In Section~\ref{sec:ref_resolution}, we then test the effectiveness of our pre-training approach for the task of reference resolution.  

\subsection{Modality Encoders}
\label{subsec:encoders}

We use three encoders
to extract representations of speech, text, and body movements, respectively.

\paragraph{Speech.} 
As speech encoder, we use multilingual wav2vec-2 (version  \texttt{wav2vec2-xlsr-300}), a masked-language model pre-trained on a large number of speech datasets in multiple languages \cite{baevski2020wav2vec, conneau2020unsupervised}. Similarly to  \citet{pepino2021emotion}, we aggregate the embeddings across all Transformer layers using a learnable weighted average and pass the output through two point-wise CNN layers to fuse signals along the temporal dimension. 

\paragraph{Semantics.} 
Although wav2vec-2 representations may capture diverse linguistic properties including prosody, phonetics, and to some extent semantics \cite{tsai-etal-2022-superb, zaiem2025speech}, they are less semantically rich than word embeddings learned from text. Therefore, we also experiment with the word embeddings 
from a pre-trained Dutch BERT-based model \cite[BERTje;][]{devries2019bertje}.

\paragraph{Skeleton.}
We adapt DSTFormer \cite{zhu2023motionbert} to encode sequences of body movements. 
The original model has two parallel branches: one applies temporal self-attention followed by spatial self-attention, and the other one spatial followed by temporal. 
To reduce overhead, in each encoder, we keep only one temporal layer and one spatial layer in each branch and replace the second layer with an optional cross-attention module. This optional cross-attention takes semantic or speech embeddings as keys and values, as schematically illustrated in Figure~\ref{fig:multimodal-x_approach}.

\subsection{Model Architectures}
\label{sec:architectures}

We propose three pre-training strategies to learn gesture representations. The first one is unimodal, in the sense that it learns representations only considering the body movements that make up a gesture. The other two (multimodal and multimodal-X) are motivated by a more holistic conception of co-speech gestures as multimodal acts \cite{HollerLevinson2019, ozyurek2014hearing}, and therefore exploit both skeletal and concurrent verbal input. We describe the gist of each architecture here and provide further technical details in Appendix~\ref{app:models}.

\paragraph{Unimodal architecture.}
This model only leverages skeletal information. It jointly optimizes a masked reconstruction loss and a unimodal contrastive loss.  For the former, we follow the original procedure for pre-training DSTFormer \cite{zhu2023motionbert} by randomly masking portions of the 2D keypoint skeletal input and learning to reconstruct them. The unimodal contrastive loss pulls representations of two views of augmented skeletal data closer while pushing them away from other negative samples in a batch.
A detailed diagram of this architecture is shown in Figure \ref{fig:unimodal_approach}, Appendix~\ref{app:models}.

\paragraph{Multimodal architecture.}
This model combines the two losses from the unimodal architecture with a \textit{multimodal} contrastive loss 
that integrates skeletal information with either speech or semantics. For the multimodal contrastive loss, we use a CLIP-like contrastive objective \cite{radford2021learning} mapping global representations of gestures and co-occurring utterances (as either raw speech or semantics) into a joint feature space.

\paragraph{Multimodal-X architecture.}
\update{Finally, this model is optimized with two complementary contrastive losses, as illustrated in Figure \ref{fig:multimodal-x_approach}: the multimodal contrastive loss  (blue arrows) described as part of the multimodal architecture above and an additional 
\textit{crossmodal} contrastive loss (green arrows), which aligns the unimodal skeleton representation with the fused gesture-semantic (or gesture-speech) embedding---the latter is obtained by injecting text tokens (or speech frames) into our DSTFormer backbone via the cross-attention layers described in Section \ref{subsec:encoders}.}

\update{The unimodal and multimodal architectures can thus be seen as ablations of the arguably more powerful multimodal-X architecture: the multimodal architecture does not include cross-attention layers, and the unimodal architecture omits multimodal alignment altogether.}



\paragraph{Training and implementation details.}
We train the three architectures above using CABB-L and CABB-XL, which allows us to test the impact of increasing the size of the training data. In Appendix \ref{app:implementation_details}, we provide further details about the backbone models, projection heads, and the parameters used in the learning objectives, along with the implementation details.

\begin{figure}
    \centering
    \includegraphics[width=1\linewidth]{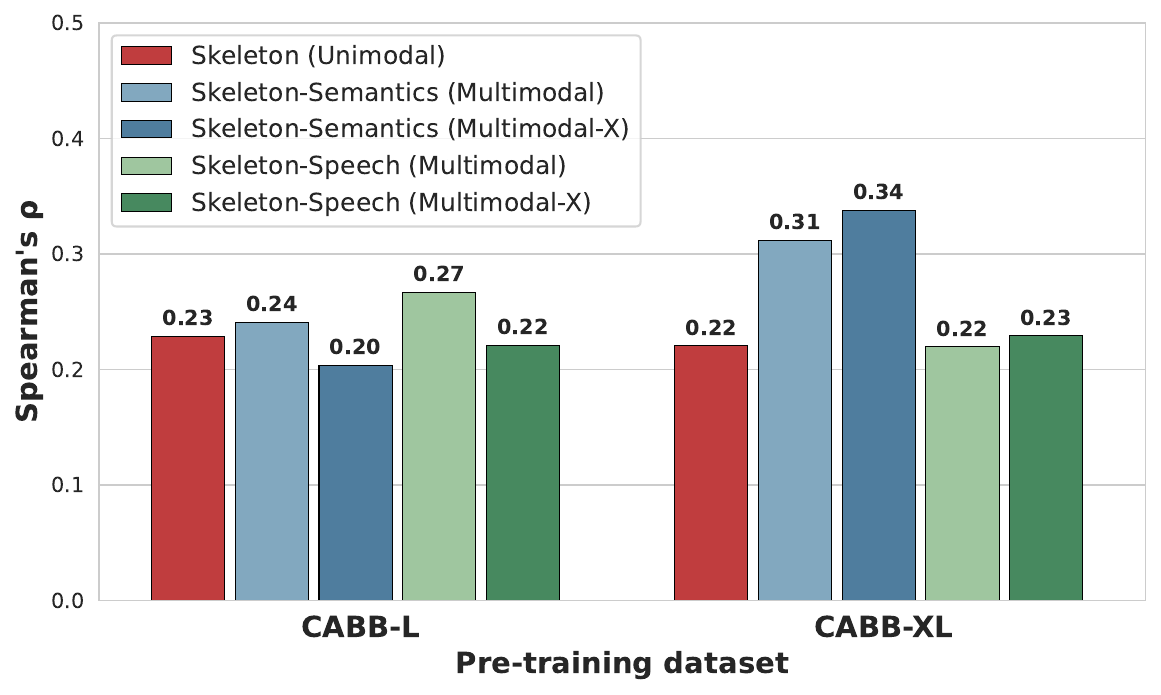}
    \caption{Spearman correlation between the number of form features shared by a pair of gestures and their cosine similarity using embeddings from skeleton-speech, skeleton-semantics, and unimodal models. Pre-training was conducted on CABB-L and CABB-XL, while the correlation scores were computed on CABB-S. All coefficients are statistically significant ($p\ll0.05$).}
    \label{fig:cabb_correlation}
\end{figure}

\subsection{Evaluation}
\label{sec:eval_pretraining}
We evaluate the gesture representations learned with our pre-training architectures using the \mbox{CABB-S} dataset, which contains manually annotated information on gestures unseen during model pre-training.
To monitor pre-training progress and save the best model variants across epochs, we conducted evaluations using form similarity as correlation.
Following \citet{ghaleb2024learning}, we compute Spearman's correlation between the number of form features a pair of gestures share according to experts' annotations and cosine similarity between the gestures' learned representations.

Figure~\ref{fig:cabb_correlation} shows the correlation results, yielded by the best models obtained during pre-training. The figure shows that the variants with the highest performance are the multimodal-X and multimodal architectures where gesture representations are jointly learned with text-based semantics from concurrent speech, using the large CABB-XL as training data. The other model variants---unimodal, multimodal (with raw speech), and multimodal-x with raw speech---do not benefit as much from an increase in the amount of training data. In fact, when exploiting raw speech, the best correlation coefficient is obtained with the multimodal architecture and CABB-L.

\begin{figure}[t]
    \centering
    \includegraphics[width=1\linewidth]{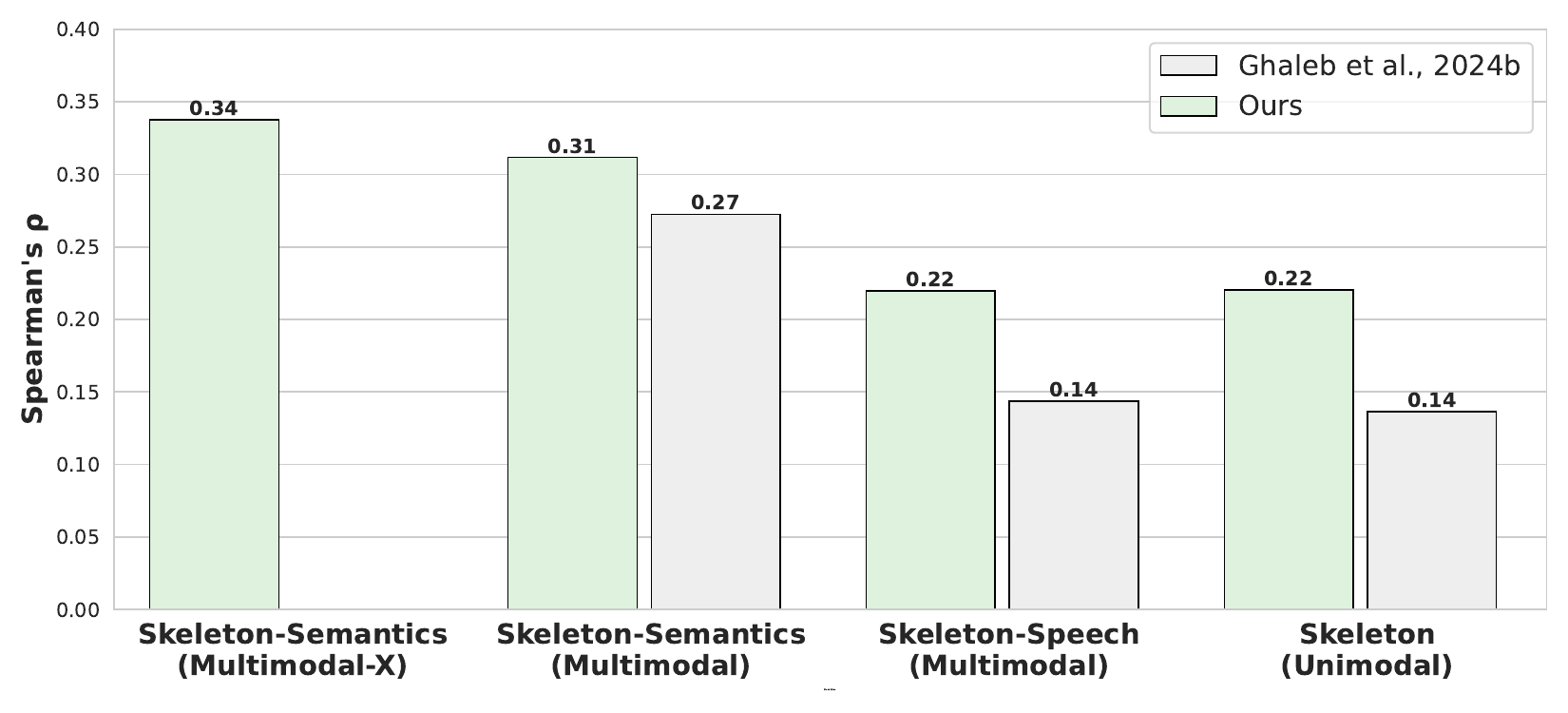}
    \caption{Pre-training on CABB-XL: comparing best models against \cite{ghaleb2024learning}. All Spearman correlation coefficients $\rho$ are statistically significant ($p\ll0.05$).}
    \label{fig:cabb-xl_baseline_corr}
\end{figure}

\paragraph{Comparison with related work.} 
We compare our models against the framework by \citet{ghaleb2024learning}. This work proposed a pre-training approach to gesture representation learning based on Spatio-Temporal Graph Convolutional Networks (ST-GCN) using unimodal and multimodal contrastive learning with co-occurring raw speech. For comparability with our approach, when reproducing this framework, in addition to raw speech we extend it to also use text-based semantics and pre-train it on CABB-XL.\footnote{Due to the architecture of ST-GCN, it is not possible to combine it with the multimodal-X architecture introduced in our work (Figure \ref{fig:multimodal-x_approach}).} 
The comparison of correlation coefficients with form similarity is shown in Figure~\ref{fig:cabb-xl_baseline_corr}. As can be observed, the gesture representations learned by our Transformer-based encoder are more aligned with form-based expert annotations as evidenced by higher correlation values across the board.

Overall, the correlation analysis indicates that the best pre-training strategies combine skeletal data with semantic information—using either multimodal or multimodal-X approaches—specifically when trained on a large dataset like CABB-XL. In the next section, we focus on model variants trained on CABB-XL that use semantic embeddings, with unimodal models serving as a baseline.

\begin{figure}
    \centering
    \includegraphics[width=1\linewidth]{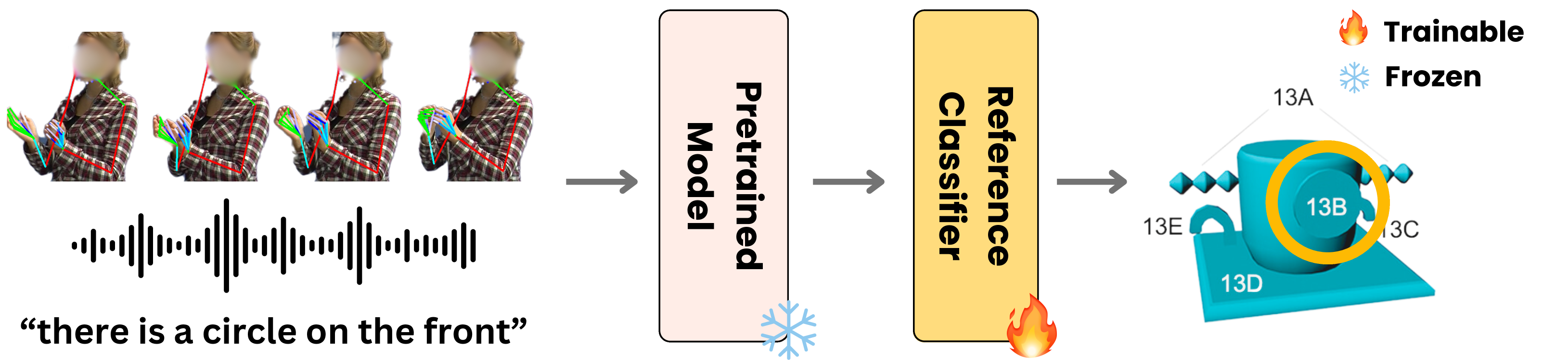}
    \caption{\update{Gesture-based reference resolution. The reference resolution classifier leverages gesture representations encoded with our pre-trained models.} 
    }  \label{fig:pretraining_and_reference_resolution}
\end{figure}

\section{Reference Resolution}
\label{sec:ref_resolution}
An important question in situated interactions is to what extent representational gestures complement or supplement speech in reference resolution. Here we shed light on this question by leveraging our pre-trained gesture models for the downstream task of reference resolution. We investigate whether models that have learned gestures by exploiting multimodal information (from body movements and concurrent speech) have more predictive power than models that represent gestures exclusively from body movements. Moreover, we test whether gestures contribute complementary information to the verbal modality when identifying referents. 


\subsection{Resolution Model and Evaluation Setup}
The resolution model leverages our model architectures pre-trained on CABB-XL without any fine-tuning. \update{This is schematically shown in Figure \ref{fig:pretraining_and_reference_resolution}}. The model is implemented as a multi‐class MLP classifier with two hidden layers of size 300 and 150, respectively, and it is trained on CABB-S. Given a gesture encoded with our pre-trained models, we train the MLP to predict one referent among 70 possible object sub‐parts (see Appendix~\ref{app:objects} for details) using a batch size of 32 and a learning rate of $10^{-4}$ with the Adam optimizer for 200 epochs. 
Recall from Section~\ref{sec:data} that each dialogue consists of six rounds. We use leave-one-round-out cross-validation, holding out the gestures in one round as a test set and training on the gestures in the remaining rounds across all dialogues in CABB-S. 
We use accuracy as an evaluation metric. 
\update{Since speakers can refer to any of 70 possible sub-parts (across the 16 objects), randomly guessing results in a rate of 1/70, leading to approximately 1.4\% accuracy.}
A model using random gesture embeddings (without access to our pre-trained models) achieves around 3\% accuracy.

\update{Given a gesture unseen during pre-training, we investigate two scenarios: 
In Section~\ref{subsect:unimodal_evals}, we measure how accurate a reference resolution system that only has access to the gesture (\ie, to skeletal input) is at predicting the gesture's referent. Here the gesture embedding is extracted zero-shot with our models, some of which exploited raw speech or text semantics during pre-training---but importantly verbal input is not provided at inference time in this scenario. In Section~\ref{subsec:resolution2}, we consider a second scenario, where at inference time the reference resolution system has access to both the unseen gesture to be resolved and any concurrent speech. We operationalise this by concatenating the gesture embedding extracted with our models and a semantic embedding of the concurrent speech, and then measure whether this leads to higher reference resolution accuracy than only exploiting the semantics of concurrent speech.}

\subsection{Gesture-Only Reference Resolution}
\label{subsect:unimodal_evals}

\begin{figure}
    \centering
    \includegraphics[width=1\linewidth]{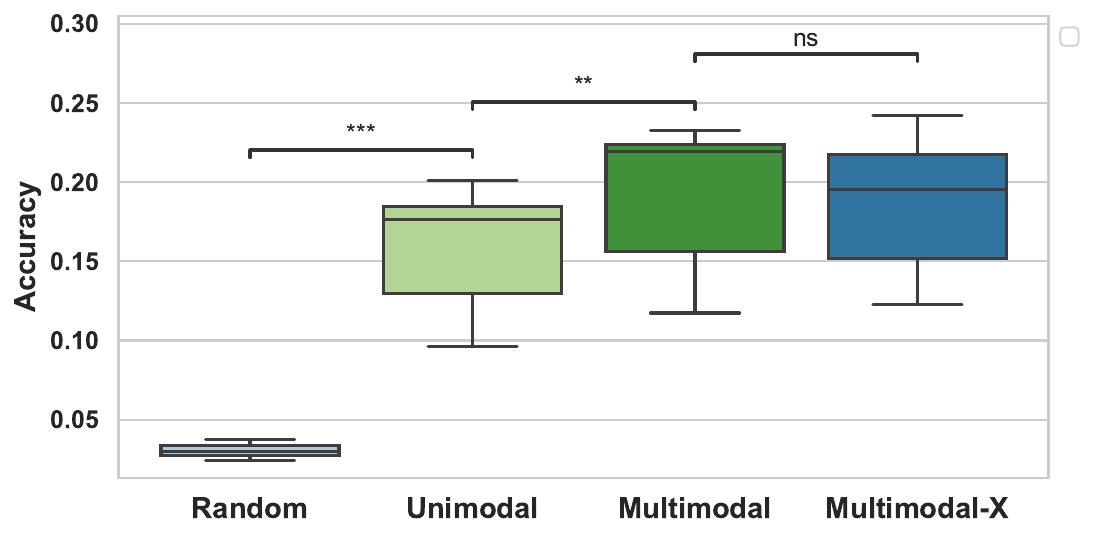}
    \caption{Average reference resolution accuracies for gesture-only embeddings are shown for unimodal, multimodal, and multimodal‑x representations. The multimodal and multimodal‑x models are pre‑trained with text‑based semantic input, and the embeddings are derived only from skeletal data. \update{Asterisks indicate statistically significant differences, with ** $p < 0.01$ and  *** $p < 0.001$; `ns' stands for `not significant'.}}
\label{fig:reference_resolution_from_gestures}
\end{figure}

We first evaluate the resolution model when it only has access to gestural information (\ie, skeletal data) as input. As shown in Figure~\ref{fig:reference_resolution_from_gestures}, 
when a gesture is encoded with our unimodal model, the average resolution accuracy is 16\%, significantly above the random baselines. 
Using embeddings from models 
that were pre-trained jointly with text-based semantics
significantly increases resolution accuracy to around 19\%, with no statistically significant difference between the multimodal and multimodal-X approaches.
These results show that our pre-trained gesture representations capture information that is useful to identify referents.\footnote{\update{In Appendix~ \ref{app:experiments_with_noisy_data}, we present a supplementary experiment showing that this is the case even in the presence of noisy skeletal data, which further emphasizes the robustness of our gesture representations.}} 
Moreover, they indicate that learning gesture representations by jointly exploiting body movements and the semantics of co-occurring speech enhances their reference resolution potential, even when information about concurrent speech is not provided at prediction time.

\begin{figure}
    \centering
    \includegraphics[width=1\linewidth]{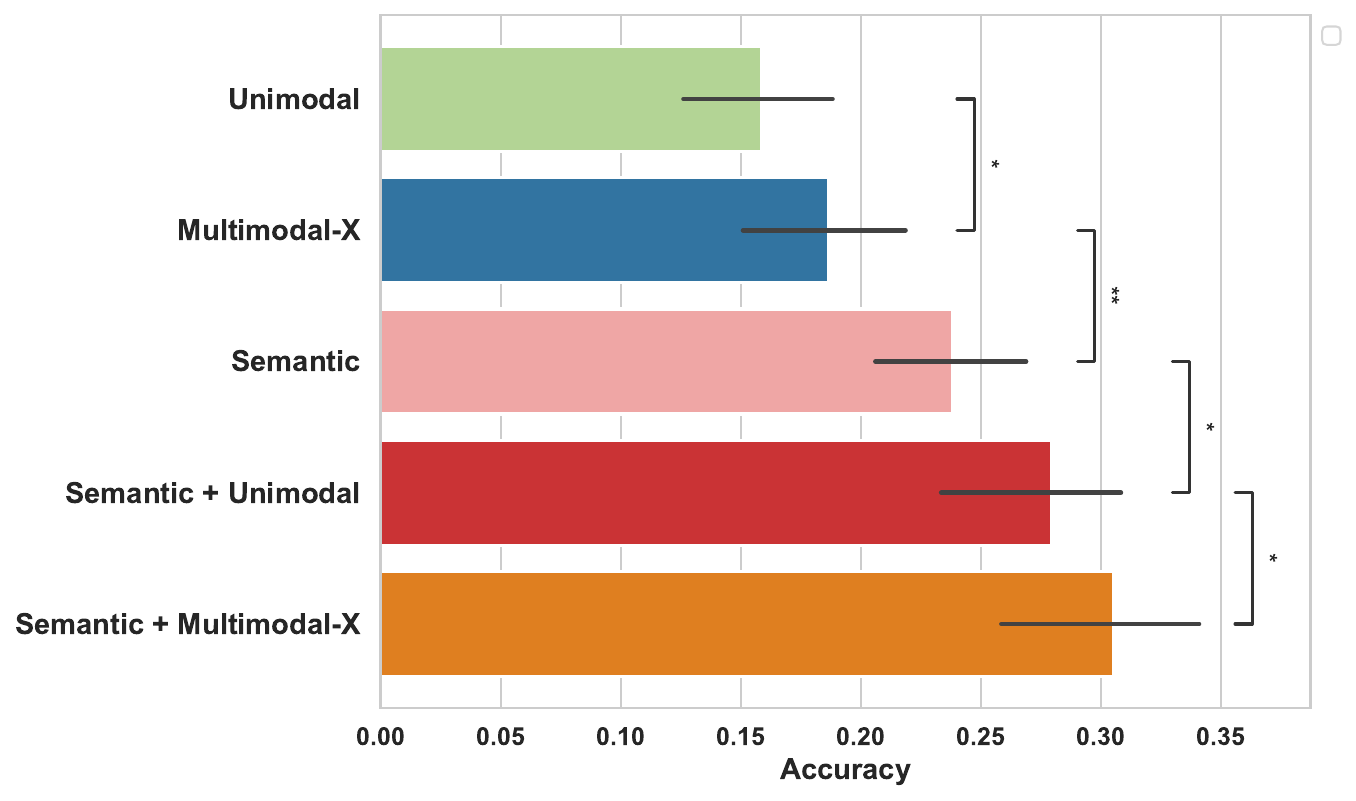}
     \caption{Average accuracies for reference resolution using gestures and co-occurring verbal information are reported for unimodal gesture embeddings, multimodal‑x gesture embeddings, semantic embeddings, and their concatenated representations. The multimodal‑x gesture embeddings are learned through pre‑training with semantic embeddings. \update{Asterisks indicate statistically significant differences, with * $p< 0.05$, ** $p< 0.01$ and *** $p< 0.001$}.}
    \label{fig:reference_resolution_with_semantic}
\end{figure}

\subsection{Reference Resolution with Gestural and Co-occurring Verbal Information}
\label{subsec:resolution2}
Next, we assess the resolution model when it has access to gestures and the speech co-occurring with them. \update{As mentioned above,} we operationalise this scenario by concatenating a gesture embedding (extracted with either unimodal or multimodal-X pre-trained models)\footnote{Given the lack of statistically significant difference between multimodal and multimodal-X in Section~\ref{subsect:unimodal_evals} Fig.~\ref{fig:reference_resolution_from_gestures}, we focus on the multimodal-X model for this experiment. } with a semantic embedding derived from the transcribed co-occurring speech using BERTje \cite{devries2019bertje}. 
The results are shown in Figure~\ref{fig:reference_resolution_with_semantic}, where we also include a condition where the reference resolution model exclusively uses the concurrent speech in the form of a semantic embedding. In that condition, resolution accuracy is 24\%. That is, the concurrent information present in the verbal modality has stronger predictive power to identify referents than body movements alone, which is to be expected in spoken conversations. 
Importantly, when both the verbal and gestural modalities are combined, we observe a significant increase in reference resolution accuracy, reaching 31\% when gestures are encoded with our multimodal-X model, pre-trained with text-based semantics. These findings confirm the complementary roles of gesture and speech in reference resolution and highlight the benefits of exploiting such complementarity for gesture representation learning.\looseness=-1

\subsection{Impact of Dialogue History}
\label{subsec:history}

It is well known that, in referential communication tasks, participants tend to reuse the same referential expressions over the course of the dialogue, creating dialogue-specific conventions \cite{clark1986referring, brennan1996conceptual}. Such `alignment' has been observed for both speech and gestures \cite{Akamine2024sp}. Hence, a system tasked with identifying the referent of a gesture is expected to achieve higher accuracy if it has access to other gestures previously used within the same dialogue than if such dialogue history is not available. 
To test whether our approach to gesture representation learning gives rise to this pattern, we train two versions of our reference resolution model: a \textit{baseline} model and a \textit{dialogue-specific} model.  The baseline model is trained on all dialogues in CABB-S, except the target dialogue---thus, referent prediction for the gestures in the target dialogue is carried out without dialogue history. In contrast, the dialogue-specific model is progressively adapted over the dialogue rounds: \ie, in round 1 it is identical to the baseline model, but 
by round $n$, it has additionally seen data from all previous dialogue rounds up to $n-1$.
To keep the amount of training data comparable between the baseline and dialogue-specific models, when new round data is added, we proportionally reduce the amount of data drawn from other dialogues during the re-training of the dialogue-specific model. As a result, both models are trained on an identical number of samples in every round.

\begin{figure}
    \centering    \includegraphics[width=1\linewidth]{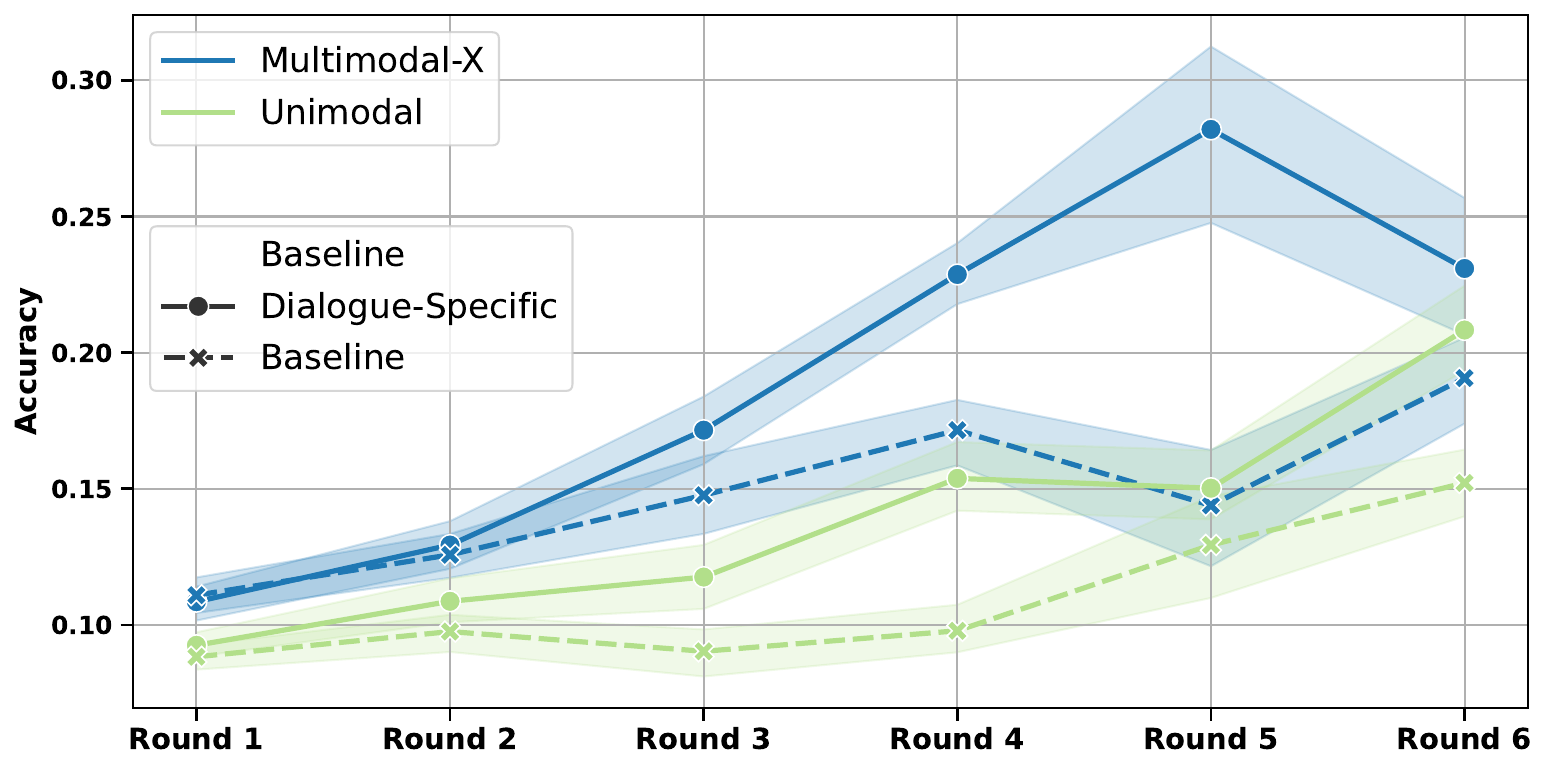}
    \caption{Average reference resolution accuracy over dialogue rounds using gesture embeddings from the unimodal (green) and multimodal-X (blue) pre-trained models. Dotted lines are used for the corresponding baseline models.}
    \label{fig:classification_with_dialogue_history}
\end{figure}

To isolate the impact of dialogue history on gesture-driven reference resolution, in this experiment we focus on identifying referents with only gestural information as input (as in Section~\ref{subsect:unimodal_evals}),\footnote{The impact of dialogue history on text-based reference resolution has already been extensively studied, \eg, \citet{shore2018using,haber2019photobook,takmaz2020refer, hawkins-etal-2020-continual}.} 
comparing our unimodal and multimodal-x pre-trained models. 
Figure~\ref{fig:classification_with_dialogue_history} shows that as the conversation unfolds over the rounds, the dialogue-specific reference resolution models outperform the baselines (dotted lines). Our statistical analysis shows that there is a significant difference in accuracy between the two (independent t-test yielding $t=2.9$, $p\ll 0.05$ for both the unimodal and multimodal-x models). It also indicates that the pattern of increased accuracy, as more dialogue history becomes available, is more pronounced when the gestures are encoded with the multimodal-x pre-trained model (Spearman correlation between accuracy values and dialogue round numbers: $\rho=0.32$ for the unimodal model and $\rho=0.35$ for the multimodal-x model, with $p\ll 0.05$ in both instances).\footnote{Note that there is no statistically significant difference in accuracy between rounds 5 and 6, despite the apparent drop.}

Overall, the results indicate that our gesture representations, particularly when learned via multimodal-x pre-training, encode features that capture the subtle increase in similarity between gestures referring to the same object within a given dialogue. In other words, to some extent the models capture gesture entrainment, which results in an advantage for the task of reference resolution.  From a practical point of view, this suggests that access to dialogue history can be an asset to agents deployed with a gesture resolution model.

\section{Conclusion}

In this work, we have studied representational co-speech gestures in collaborative dialogue, using an existing dataset of face-to-face interactions collected by cognitive scientists. We introduced a novel reference resolution task formulated as the problem of identifying the intended referent of a co-speech gesture, while addressing key challenges in gesture representation learning. We proposed a self-supervised Transformer-based approach to learning pre-trained gesture embeddings by jointly exploiting skeletal information and concurrent language encoded with text or speech large language models. Our experiments showed that the resulting gesture embeddings effectively contribute to reference resolution. Representing gestures by exclusively exploiting skeletal information has significant predictive power, and grounding body movements in concurrent speech during pre-training further improves resolution accuracy, even when speech is not provided at test time. 
\update{An interesting avenue for future work could be to ground the reference resolution models in the visual properties of the referents, in order to learn mappings between iconic gestures and the objects they represent, which might lead to further improvements.}
Moreover, we showed that reference resolution from representational gestures can benefit from having access to gestures previously used within a dialogue, thus providing empirical support for the presence of gestural entrainment in face-to-face interaction.

Taken together, our findings emphasize the multimodal character of conversation \cite{HollerLevinson2019, ozyurek2014hearing} and the importance of capturing the complementarity between gestures and speech in naturalistic human-machine interaction. 
Further work is needed to test the extent to which the proposed pre-training approach would transfer to other referential domains and other tasks---a step we leave to future research.

\section*{Limitations}
The current work focuses on Dutch-speaking task-oriented dialogues, thus contributing to linguistic diversity in the current English-centric NLP landscape. We nevertheless acknowledge that it is an open question how well the proposed models may generalise to other languages, cultural contexts, tasks, as well as open-domain dialogues. On the methodological front, while we employ and adapt a state-of-the-art motion encoder and show that our pre-training objectives and architecture choices are effective, further optimisation and integration with more advanced speech and semantic encoders may give additional improvements.
\update{Finally, our method is agnostic as to whether concurrent speech and gesture are semantically congruent (i.e., express compatible content). We leverage both information streams and observe that this yields stronger representations and higher performance. Although the nature of the collaborative referential task in the CABB datasets makes it likely that the two modalities align in content, whether the performance gains stem from true cross-modal congruence remains an open empirical question that could be explored in the future. This, however,  requires manual annotation of the linguistic referential expressions, which is currently not available.}

\section*{Acknowledgements}
We thank the members of the Dialogue Modelling Group at the ILLC, University of Amsterdam, and the Multimodal Language Department at the Max Planck Institute for Psycholinguistics for their feedback. This project was partially carried out while Esam Ghaleb was at the University of Amsterdam, funded by the Dutch Research Council (NWO) through a Gravitation grant (024.001.006) to the Language in Interaction consortium. 
Raquel Fern\'{a}ndez acknowledges support from the European Research Council, ERC Consolidator Grant No.~819455.

\bibliography{custom}

\appendix
\section*{Appendix}

\section{Objects in the CABB Dataset}
\label{app:objects}
\begin{figure*}
    \centering
    \includegraphics[width=0.80\linewidth]{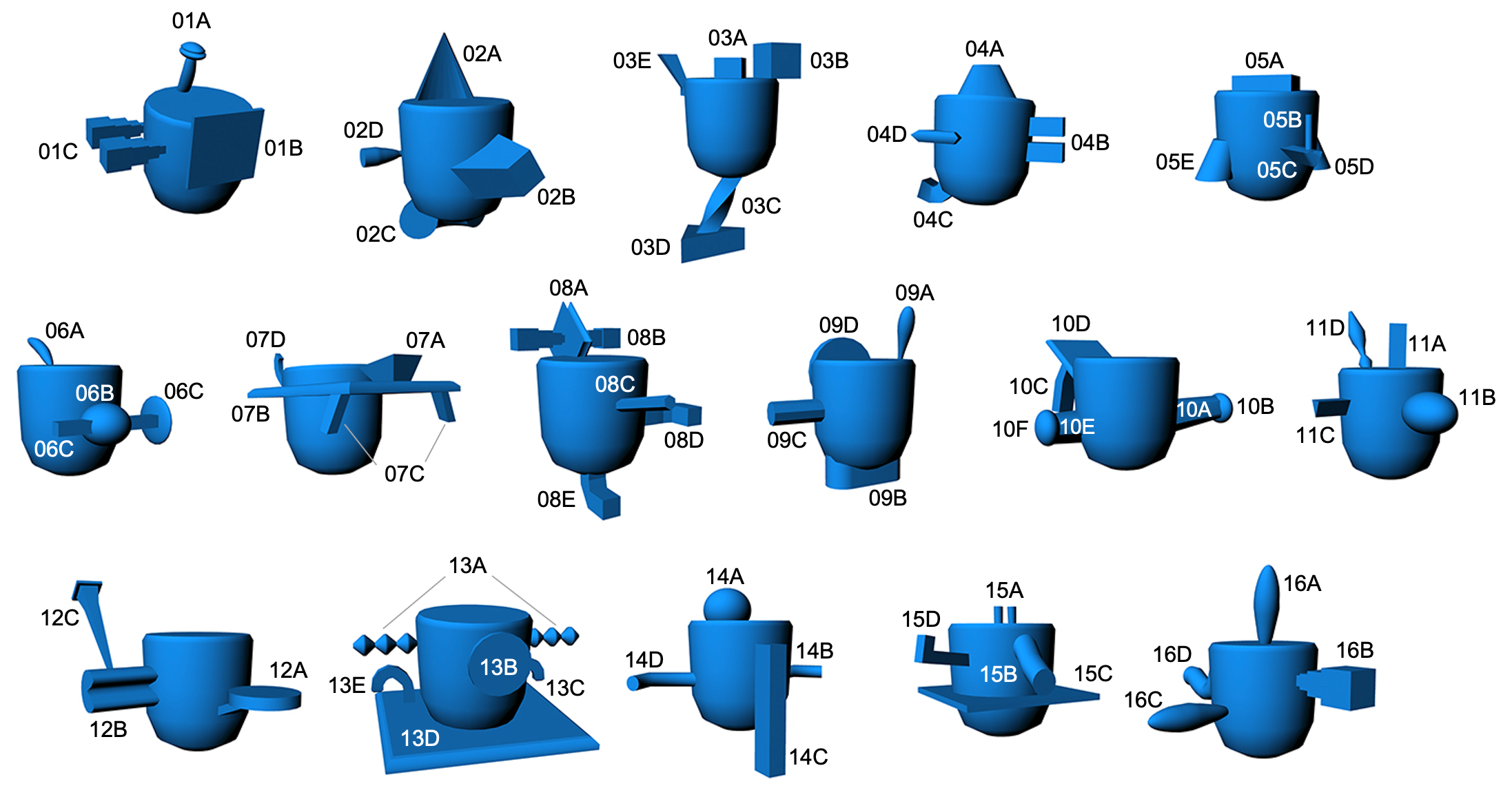}
    \caption{The candidate objects and their sub-parts present in the CABB dataset \cite{eijk2022cabb}.}
    \label{fig:all_object_and_subparts}
\end{figure*}

\begin{figure}
    \centering
    \includegraphics[width=1\linewidth]{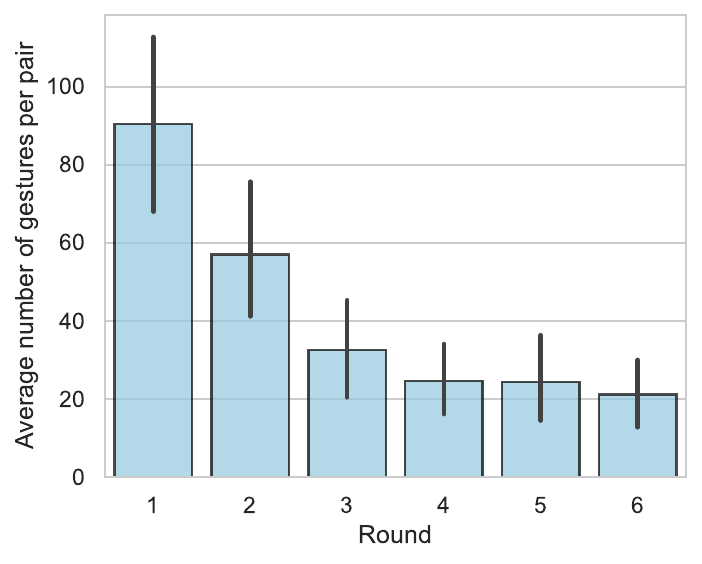}
    \caption{Distribution of manually segmented gestures across rounds of interaction.}
    \label{fig:gesture_distribution}
\end{figure}

\noindent \citet{rasenberg2022primacy} segmented each gesture stroke in the CABB-S dataset and classified them into four categories: \emph{iconic} (depicting an aspect of the target object), \emph{deictic} (explicitly pointing with an extended finger or hand), \emph{other} (e.g., beat gestures or pragmatic signals like “you go ahead”), and \emph{undecided} when the gesture type was unclear. In our study, we focus exclusively on the iconic gestures, which refer to specific parts of the novel objects in the CABB setup (see Figure \ref{fig:all_object_and_subparts}).
The average number of gestures in each round per dialogue is shown in Figure \ref{fig:gesture_distribution}.
Each iconic gesture was annotated with a sub-part label---such as \texttt{06A} for a single sub-part or \texttt{06A+06B} for multiple sub-parts. When a gesture was annotated with multiple parts, we split it into separate samples corresponding to each sub-part. Additionally, a \texttt{main} label was assigned if the gesture targeted the object’s main part, and \texttt{general} is used when the gesture indicates a broad area (e.g., “the left side”).

\section{Segmentation Qualitative Results}
\label{app:segmentation_qualitative_results}

\begin{figure}[th]
    \centering
    \includegraphics[width=1\linewidth]{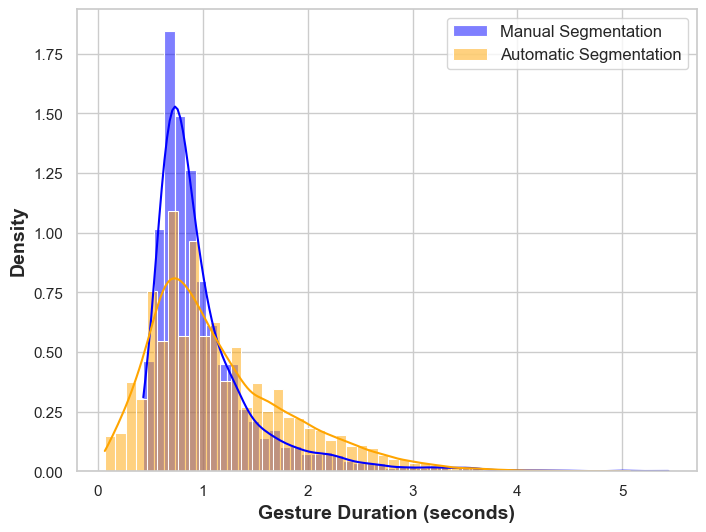}
    \caption{Durations distribution of manually and automatically segmented gestures.}
    \label{fig:gestures_durations}
\end{figure}

\begin{figure}
    \centering
    \includegraphics[width=1\linewidth]{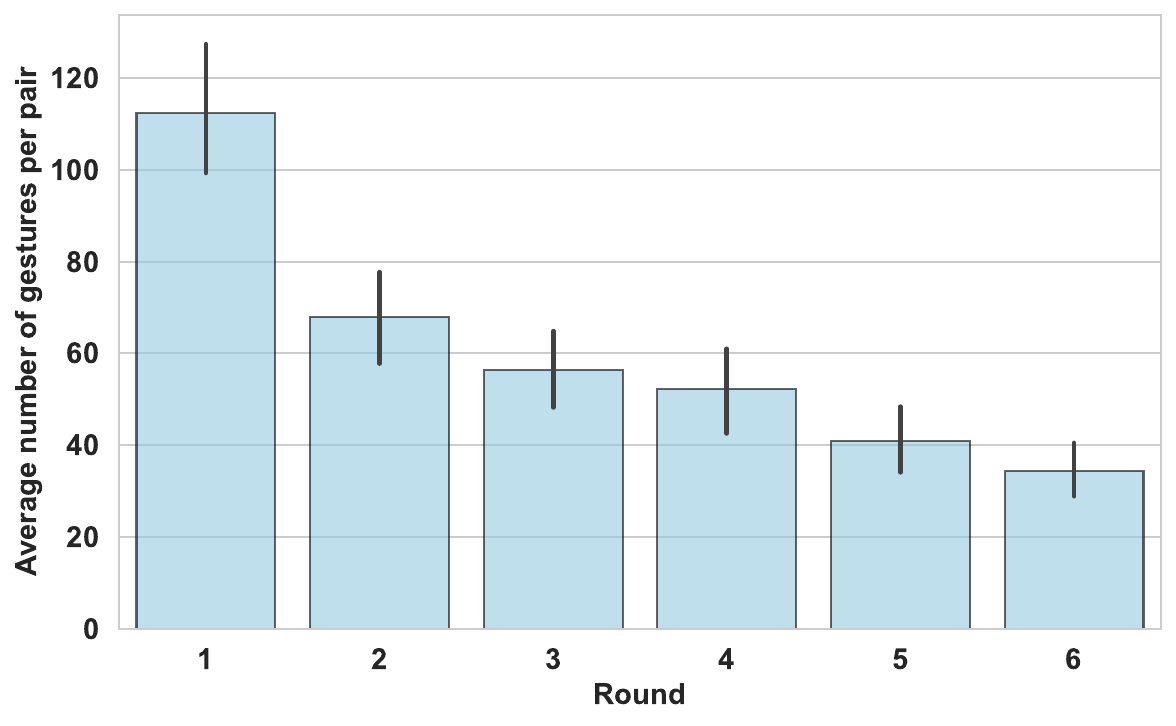}
    \caption{Distribution of automatically segmented gestures across rounds of interaction.}
    \label{fig:gesture_distribution_over_rounds}
\end{figure}

\begin{figure}
    \centering
    \includegraphics[width=1\linewidth]{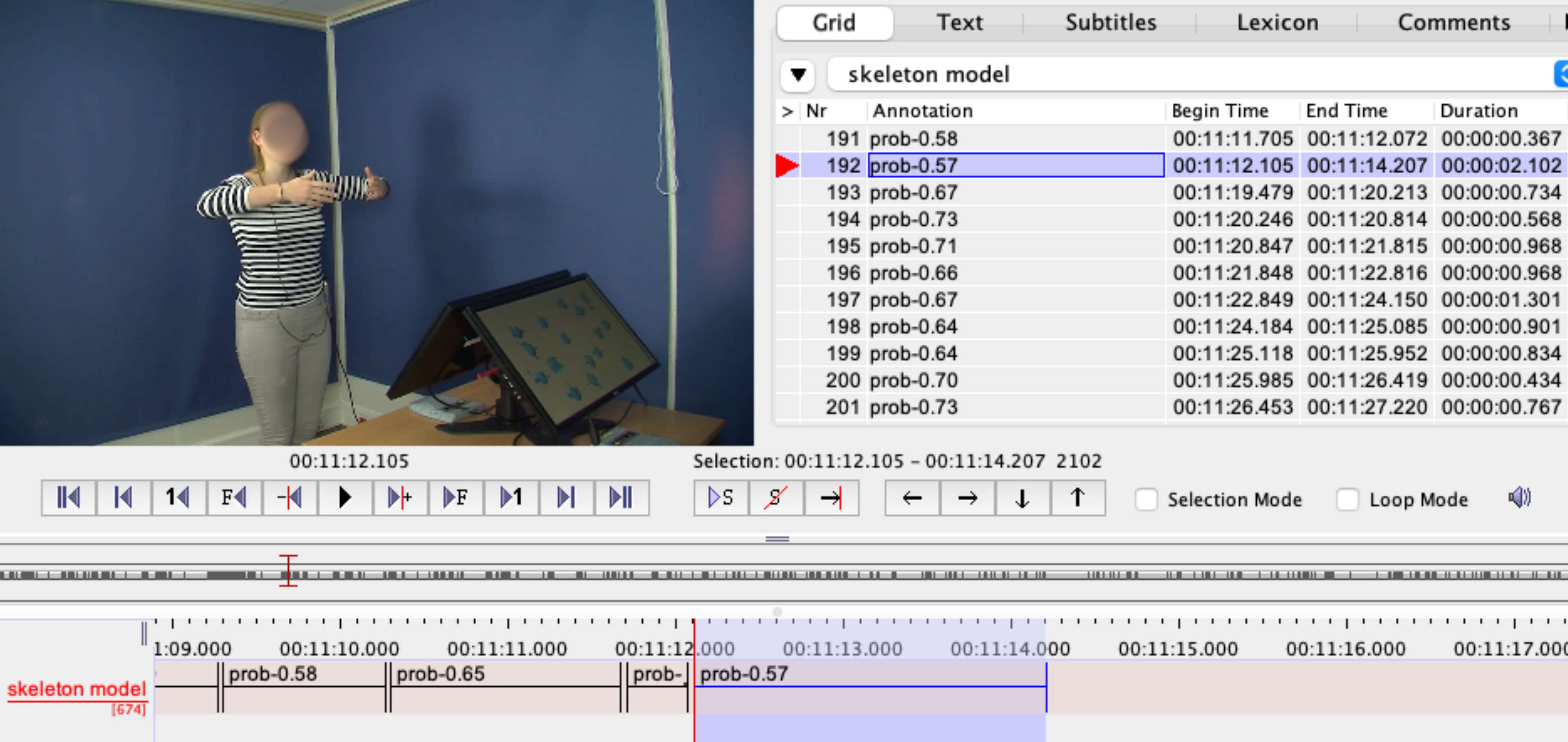}
    \caption{Screenshot of ELAN \cite{elan2024}, a popular annotation software, which we used to inspect the quality of automatic segmentation.}

    \label{fig:elan_example}
\end{figure}

\update{Figure \ref{fig:gestures_durations} compares the duration distributions of manually segmented gestures in CABB-S with those automatically segmented in CABB-L. We can see that both distribution curves peak between 0.8 and 0.9 seconds and share a right skew. The automatic segmentation shows a small portion of gestures lasting longer than two seconds. This is most likely because the segmentation model was trained on extended segments to account for the preparation and retraction phases of gestures.
Figure \ref{fig:gesture_distribution_over_rounds} plots the average number of gestures per pair across the six rounds of the referential game. Consistent with Figure \ref{fig:gesture_distribution}, gesture frequency is highest in the initial rounds and declines in later ones. 
The higher gesture count in CABB-L reflects longer interaction time: besides the referential interaction, these pairs also carried out an on-screen object localisation task that required identifying the object's position with respect to the others.
We also imported the automatically segmented gestures into ELAN \cite{elan2024} for quality check. As shown in Figure \ref{fig:elan_example}, we could visually inspect and verify that gestures' onsets were detected and segmented with high reliability. 
}

\section{Model Details}
\label{app:models}

\subsection{Pre-training objectives}
In Section \ref{sec:models}, we introduced three pre-training architectures, each containing a combination of self-supervised learning objectives. Here, we provide a detailed technical overview of these losses. The proposed architectures exploit three modalities, namely 2D skeletal keypoints and joint prediction confidence scores $\boldsymbol{X}_i^g \in \mathbb{R}^{T_g \times 27 \times 3}$ for gestures, text-based semantics  $\boldsymbol{X}_i^t$, and raw speech signals $\boldsymbol{X}_i^s$. These inputs are encoded using one of the following encoders: our adaptation of DSTFormer $f_{\Theta_g}(\cdot)$ for skeletons (Section \ref{subsec:encoders}; \citet{zhu2023motionbert}), Dutch BERT-based model $f_{\Theta_t}(\cdot)$ for text \cite{devries2019bertje}, and \texttt{wav2vec2-xlsr-300} $f_{\Theta_s}(\cdot)$ for raw speech \cite{conneau2020unsupervised}.
\label{subsec:objectives}

\begin{figure*}
    \centering
    \includegraphics[width=0.8\linewidth]{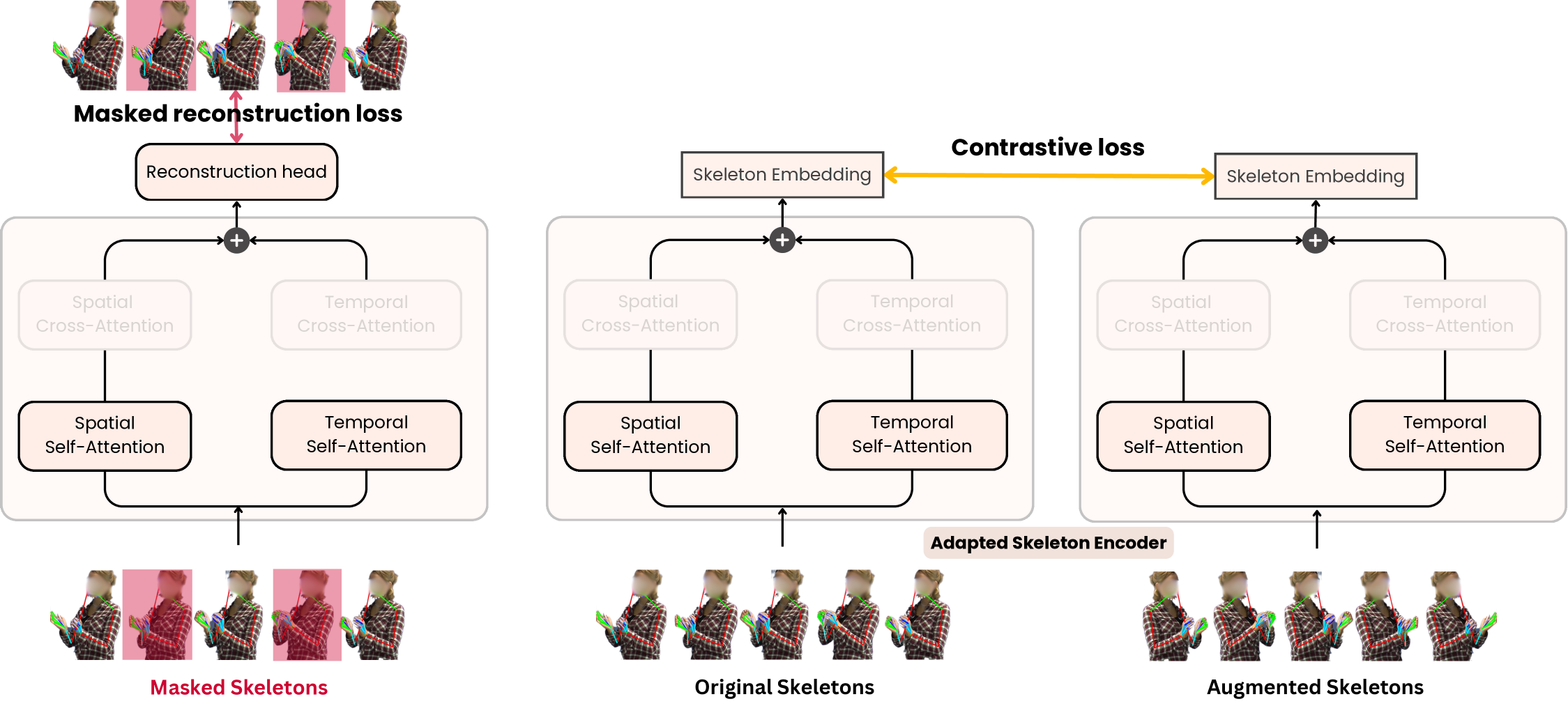}
    \caption{Unimodal architecture jointly optimizes masked reconstruction (left branch) and unimodal contrastive losses (middle and right branches). Note that cross-attention blocks are not utilized in this pre-training approach.}
    \label{fig:unimodal_approach}
\end{figure*}

\paragraph{Unimodal masked reconstruction loss.} 
We follow the original DSTFormer \cite{zhu2023motionbert} by randomly masking portions of the 2D keypoint input and learning to reconstruct them. Specifically, we accumulate a reconstruction loss between the original and predicted coordinates of masked keypoints  as follows:
\begin{equation} 
    \mathcal{L}_{k} = \sum_{t=1}^{T}  \sum_{j=1}^{J}  \delta_{t,j} ||\boldsymbol{\hat{x}}_{t,j} - \boldsymbol{x}_{t,j}||^2,
\end{equation}
where $\boldsymbol{\hat{x}}_{t,j} \in \mathbb{R}^2$ is the predicted coordinates of keypoint $j$ at timestep $t$, $\boldsymbol{x}_{t,j}$ is the ground truth keypoint, and $\delta_{t,j}$ is a weighting factor that accounts for confidence or visibility of the keypoints. 
To enforce spatial and temporal consistency, we introduce two additional reconstruction objectives $\mathcal{L}_{b}$ and $\mathcal{L}_{m}$: bone and motion reconstruction losses. The former ensures structural consistency by preserving the distances between adjacent keypoints $||\boldsymbol{x}_{t,j} - \boldsymbol{x}_{t,j-1}||$ across frames, while the latter minimizes the difference between the temporal displacement $||\boldsymbol{x}_{t,j} - \boldsymbol{x}_{t-1,j}||$ of predicted and ground truth keypoints. The overall objective for masked reconstruction is given by the average of $\mathcal{L}_{k}$, $\mathcal{L}_{b}$ and $\mathcal{L}_{m}$. Figure \ref{fig:unimodal_approach} illustrates how this objective is integrated into the unimodal pre-training architecture.

\paragraph{Unimodal contrastive loss.}
A unimodal contrastive loss is applied to different views of the same skeletal keypoint sequence distorted with simple augmentations, as illustrated in the middle and right branches of Figure \ref{fig:unimodal_approach}. Formally, for input skeleton $\boldsymbol{X}_i^g$, we obtain two augmented views $a(\boldsymbol{X}_i^g)$ and $a'(\boldsymbol{X}_i^g)$. These views are then passed through skeleton encoder $f_{\Theta_g}(\cdot)$, namely DSTFormer, and projection layers $g_{\Theta_g}(\cdot)$ to obtain projected features $\boldsymbol{z}_i^g = g_{\Theta_g}(f_{\Theta_g}(a(\boldsymbol{X}_i^g)))$ and $\boldsymbol{z}_j^g = g_{\Theta_g}(f_{\Theta_g}(a'(\boldsymbol{X}_i^g)))$. These representations are treated as a positive pair in a contrastive loss function, whereas all other views from a training mini-batch are considered negative \cite{chen2020simple}:

\begin{equation}
    l_{um}(i, j) = -log\frac{exp(\frac{s(\boldsymbol{z}^g_i, \boldsymbol{z}^g_j)}{\tau})}{\sum_{k=1}^{2b} \mathbb{I}_{[k \neq i]} exp(\frac{s(\boldsymbol{z}^g_i, \boldsymbol{z}^g_k)}{\tau})}.
    \label{eq:unimodal_contrastive_loss}
\end{equation}

\noindent
The loss maximizes cosine similarity $s(\cdot)$ for the positive pair and minimizes similarity with other views in a mini-batch of size $b$.

\paragraph{Multimodal contrastive loss.}

We propose a CLIP-like contrastive objective (depicted with a blue line in Figure \ref{fig:multimodal-x_approach}) mapping global representations of skeletons and co-occurring utterances in a joint feature space \cite{radford2021learning}. In detail, given projected representations of skeletons $\boldsymbol{z}_i^g = g_{\Theta_g}(f_{\Theta_g}(\boldsymbol{X}_i^g))$ and co-occurring utterances (\eg, text-based semantics) $\boldsymbol{z}_i^t = g_{\Theta_t}(f_{\Theta_t}(\boldsymbol{X}_i^t))$, the multimodal objective aims to maximize their similarity as follows:

\begin{equation}
    l^{g \rightarrow t}_{mm}(i)^ = -log\frac{exp(\frac{s(\boldsymbol{z}^g_i, \boldsymbol{z}^t_i)}{\tau})}{\sum_{k=1}^{b} exp(\frac{s(\boldsymbol{z}^g_i, \boldsymbol{z}^t_k)}{\tau})}.
    \label{eq:mm_loss}
\end{equation}

\noindent
The final error function accumulates losses $l^{g \rightarrow t}_{mm}$ and $l^{t \rightarrow g}_{mm}$ for each skeleton-utterance pair in a mini-batch.

\paragraph{Multimodal-X losses.}
Two losses are employed to optimize the multimodal-X architecture (Figure~\ref{fig:multimodal-x_approach}). First, the contrastive loss is computed between skeleton representations and pooled semantic embeddings in line with Equation \ref{eq:mm_loss}. Furthermore, we introduce an objective that leverages cross-attention in our adapted DSTFormer (Section~\ref{subsec:encoders}). Specifically, the representations of skeletons in one branch of the architecture are fused with semantic or speech embeddings via cross-attention layers (middle branch in Figure~\ref{fig:multimodal-x_approach}), while the other branch remains unimodal (right branch in the figure). We then apply the same contrastive formulation (Equation~\ref{eq:unimodal_contrastive_loss}) to align unimodal skeleton representations with the fused skeleton--crossmodal embeddings. This strategy encourages both robust unimodal representations and cross-modal alignment.

\subsection{Handling mismatched temporal resolutions.}
\update{Multimodal data has different resolutions. Skeletal input and speech raw waveform come with high-frequency motion frames, whereas linguistic information is tokenised into subword units of much coarser granularity.
We experiment with two integration strategies.
\begin{enumerate}
    \item \textbf{Multimodal (mean-pool).}  
    All subword embeddings in an utterance or speech waveform representations of 25 milliseconds are mean-pooled into a single vector.
    Likewise, the $T$ gesture-frame embeddings are concatenated and projected down to one vector.  
    The two global representations are then fused by the cross-modal Transformer.
    \item \textbf{Multimodal-X (frame-wise attention).}  
    Each gesture frame ($\mathbf{q}_t$) attends to the set of subword embeddings ($\mathbf{k}_j,\mathbf{v}_j$) for the co-occurring utterance through multi-head attention. We apply the same mechanisms when we handle speech frame-level representations, which operate at segments of 25 milliseconds.  
    This enables more fine-grained alignment, as each frame can focus on the specific linguistic context it co-occurs with.
\end{enumerate}
}

\begin{table*}[ht!]
\centering \small
\begin{tabular}{l|c|cc|cc}
\toprule
 & \textbf{Unimodal} & \multicolumn{2}{c|}{\textbf{Multimodal-X}} & \multicolumn{2}{c}{\textbf{Multimodal}} \\
\cmidrule{2-6}
 &  & \textbf{Semantic} & \textbf{Speech} & \textbf{Semantic} & \textbf{Speech} \\
\midrule
\shortstack[l]{\textbf{Skeleton Encoders params} \\ \textbf{+ Projection heads (M)}} 
                   & 10.3 & 22.0 & 24.2 & 10.5 & 10.5 \\
\textit{Non-trainable params (M)}   & -- & 109 & 315 & 109 & 315 \\
\textit{Total params (M)}           & 10.3 & 131 & 339 & 119 & 325 \\
\textbf{Model size (GB)}            & 0.041 & 0.525 & 1.359 & 0.478 & 1.304 \\
\bottomrule
\end{tabular}
\caption{Parameters for the three architectures. The skeleton encoders (the adapted DSTFormer) and the projection heads are the trainable parameters. The speech (i.e., wav2vec2) and text-based semantic (i.e., BERTje) encoders are frozen during pre-training.}
\label{tab:parameters}
\end{table*}

\section{Implementation Details}
\label{app:implementation_details}
All models are implemented using PyTorch \cite{Ansel_PyTorch_2_Faster_2024} and PyTorch-Lightning \cite{Falcon_PyTorch_Lightning_2019}. Training is performed on nodes with four NVIDIA A100-SXM4-40GB GPUs. The experimented three model types---unimodal, multimodal, and multimodal-X---each trained using its respective objective for a maximum of 100~epochs. We use Adam optimizer with a learning rate of 0.001. For multimodal-X, a per-GPU batch size of 96 (for a total of 384 across four GPUs) strikes a balance between VRAM utilization and achieving reliable convergence. For multimodal models, we could only fit a batch size of 64 per GPU due to using the additional model for masked reconstruction. We also used a batch size of 64 for the unimodal models. 
For the contrastive objective, we set the temperature to 0.1 by default. Masked reconstruction follows the DSTFormer \cite{zhu2023motionbert} configuration with a masking probability of 5\% and an equivalent amount of noise applied. We randomly split 90\% of our generated time windows in CABB-XL for pre-training and use the remaining 10\% for validation. The CABB-S dataset is reserved solely to monitor pre-training convergence and agreement with expert annotations; we select final checkpoints based on these performance metrics. Throughout model training, we employ data augmentations similar to those proposed by \citet{ghaleb2024an}. Namely, we apply various skeletal transformations (mirror, shift, scaling, rotation, jitter, shear) to ensure the models generalize to pose variability.

\subsection{Implementation of the skeleton encoder}
\label{app:skeleton_encoder}
The adapted DSTFormer encoder processes skeletal data with two parallel branches that separately attend to spatial and temporal features.  In its unimodal version (without cross-attention), the encoder consists of 4 blocks per branch (8 blocks overall), each containing standard attention and MLP layers with residual connections. The encoder’s output is then fed into a projection head---an MLP that maps the encoded features (e.g., from 256 to 128 dimensions)---to produce the final feature representation.

In the multimodal-x variant, each block contains an additional cross-attention module that fuses either text-based semantic embeddings or speech embeddings with the skeleton representation. This extension doubles the number of trainable parameters, resulting in about 22.0 million parameters when using semantic inputs and 24.2 million when using speech inputs. The difference in parameter counts is due to the different embedding sizes from the semantic (768 dimensions) and speech (1024 dimensions) backbones and the subsequent projection heads. Similar to the unimodal case, dedicated projection heads then map the features to the shared embedding space. The multimodal‑x model pre‑trained with text-based semantics takes approximately 15 hours to run, while the multimodal model runs for roughly 12 hours. The unimodal one requires considerably less time (around 8 hours for 100 epochs) since it does not rely on the backbone models of co-occurring speech. 

In Table \ref{tab:parameters}, we summarize the number of parameters for each model architecture.

\section{Impact of Errors on Pose Estimation}
\label{app:experiments_with_noisy_data}

\begin{table}[t]
\centering \small
    \begin{tabular}{@{}lcc@{}}
        \toprule
        \bf Model variant & $\sigma$ & Accuracy (mean $\pm$ s.d.) \\
        \midrule
        Multimodal-X (clean)        & 0.00 & 0.19 $\pm$ 0.05 \\
        Multimodal-X (jittered)     & 0.20 & 0.18 $\pm$ 0.04 \\
                                    & 1.00 & 0.19 $\pm$ 0.05 \\
                                    & 15.0 & 0.16 $\pm$ 0.04 \\
        \bottomrule
    \end{tabular}
    \caption{Reference resolution accuracy using the multimodal-X architecture with skeletal information including different degrees of noise (jittered).}
    \label{tab:pose_noise}
\end{table}

\update{There may be errors in pose estimation because we rely on off-the-shelf ViTPose~\cite{xu2022vitpose} and, in the CABB dataset set-up, some joints are occasionally occluded.
Several design choices were therefore made to make the models' representations robust.
First, we employ self-supervised objectives, which are known to decrease the effect of noisy inputs \cite{hendrycks2019using}.
Second, we feed the per-joint confidence scores returned by ViTPose as an additional channel in the skeletal input (see Appendix \ref{app:models}), so the model can learn to down-weight unreliable joints.
\paragraph{Additional noise experiment.}
To test the models’ resilience to noisy data, we conduct an additional experiment by adding Gaussian noise with varying jitter ($\sigma$) into the 2D skeleton coordinates. Table~\ref{tab:pose_noise}  shows the reference resolution accuracies achieved by the skeleton encoder pre-trained with multimodal-X when different degrees of noise are added. The results show that our model is reasonably robust and only at very high noise levels ($\sigma > 10$, which exceeds the typical error rate of current pose estimators) does performance drop slightly.}



\end{document}